\documentclass[journal]{IEEEtran}
\usepackage{amsmath,graphicx,color}
\usepackage{array}
\usepackage{longtable}
\usepackage{cite,amssymb}
\usepackage{subcaption}
\usepackage{authblk}
\usepackage{dblfloatfix}

\usepackage{url}
\usepackage{soul}
\usepackage{graphicx,color}
\usepackage[dvipsnames]{xcolor}
\newcommand*{\RR}[1]{\textcolor{Plum}{#1}}
\newcommand*{\SL}[1]{\textcolor{red}{#1}}

\usepackage[english]{babel}
\usepackage{lipsum}
\usepackage[pscoord]{eso-pic}

\newcommand{\placetextbox}[3]{
  \setbox0=\hbox{#3}
  \AddToShipoutPictureFG*{
    \put(\LenToUnit{#1\paperwidth},\LenToUnit{#2\paperheight}){\vtop{{\null}\makebox[0pt][c]{#3}}}%
  }%
}%
\begin{document}


\placetextbox{0.5}{1}{\SL{****\scriptsize{This work has been submitted to the IEEE Sensors J. for possible publication. Copyright may be transferred without notice, after which this version may no longer be accessible.}****}}
\title{Phonocardiographic Sensing using Deep Learning for Abnormal Heartbeat Detection}

\author[1]{Siddique Latif}
\author[2]{Muhammad Usman}
\author[3]{Rajib Rana}
\author[1]{Junaid Qadir}

\affil[1]{Information Technology University (ITU)-Punjab, Pakistan}
\affil[2]{COMSATS Institute of Information Technology, Islamabad}
\affil[3]{University of Southern Queensland, Australia}

\maketitle

\begin{abstract}

Cardiac auscultation involves expert interpretation of abnormalities in heart sounds using stethoscope. Deep learning based cardiac auscultation is of significant interest to the healthcare community as it can help reducing the burden of manual auscultation with automated detection of abnormal heartbeats. However, the problem of automatic cardiac auscultation is complicated due to the requirement of reliability and high accuracy, and due to {the presence of background noise in the heartbeat sound}. In this work, we propose a Recurrent Neural Networks (RNNs) based automated cardiac auscultation solution. Our choice of RNNs is motivated by the great success of deep learning in medical applications and by the observation that RNNs represent the deep learning configuration most suitable for dealing with sequential or temporal data even in the presence of noise.  We explore the use of various RNN models, and demonstrate that these models deliver the abnormal heartbeat classification score with significant improvement. 
Our proposed approach using RNNs can be potentially be used for real-time abnormal heartbeat detection in the Internet of Medical Things for remote monitoring applications.

\end{abstract}

\IEEEpeerreviewmaketitle

\section{Introduction}

Cardiovascular diseases (CVDs) are the major health problem and have been the leading cause of death globally. They are causing nearly 48\% deaths in Europe \cite{allender2008european}, 34.3\% in America \cite{lloyd2010heart}, and more than 75\% in developing countries \cite{latif2018mobile}. Earlier diagnosis of CVDs is crucial as it can drastically decrease the potential risk factors of these deaths\cite{yang2011prevalence}. 

Auscultation is an widely used method for CVD diagnosis, which uses stethoscope to find out clues of different cardiac abnormalities. However, it requires extensive training and experience for proper auscultating \cite{roy2002helping}. Only 20 to 40\% accuracy can be achieved when performed by medical students and primary care physicians \cite{mangione1997cardiac,lam2005factors}, and roughly 80\% can be achieved when conducted by expert cardiologists \cite{strunic2007detection,ejaz2004heart}. 

Another effective solution is echocardiograms that visualize the heart beating and blood pumping. However, this procedure is expensive with an average cost of \$1500 as per current cost \cite{cost}. Overall, there is a distinctive lack of a reliable yet cost-effective tool {for earlier diagnosis of CVD.}

A potential solution to the above-mentioned problems is a sensor-enabled automated diagnosis in real-time using mobile devices and cloud computing \cite{springer2015mobile,latif20175g}. In fact, this has inspired the development of phonocardiography (PCG), which is now an effective and non-invasive method for earlier detection of cardiac abnormality \cite{ari2009search}. In PCG, heart sound is recorded from the chest wall using a digital stethoscope and this sound is analyzed to detect whether the heart is functioning normally or the patient should be referred to an expert for further diagnosis. 
Due to its high potential, automatic detection of cardiac abnormalities through PCG signal is a rapidly growing field of research \cite{maglogiannis2009support}.
Most of these automated cardiac auscultation attempts, however, have utilized either classical machine learning models (e.g., \cite{wang2007phonocardiographic,saraccouglu2012hidden,avendano2010feature,zheng2015novel}) or feed-forward neural networks \cite{uuguz2012biomedical,nassralla2017classification} rather than Recurrent Neural Networks (RNNs)---which are intrinsically better suited for this task. 

Heart sound is a physiologic time series and it possesses temporal dynamics that change based on the different heart symptoms. Due to their well-known capabilities for modeling and analyzing sequential data even in the presence of noise \cite{hochreiter1997long}, in this paper we explore various RNNs for automated cardiac auscultation. 
We use 2016 PhysioNet Computing in Cardiology Challenge dataset \cite{liu2016open} that contains 
phonocardiograms recorded using  sensors placed at the four common locations of human body: pulmonic area,  aortic area, mitral area, and tricuspid area.  The work presented in this study is the \textit{first attempt} that investigates the performance of various state-of-the-art RNNs for heartbeat classification using PCG signals.


\section{Related Work}

In the past few years, automatic analysis of phonocardiogram (PCG), which refers to the high-fidelity sound recording of murmurs made by the heart, has been widely studied especially for automated heartbeat segmentation and classification.  According to Liu et al. \cite{liu2016open} there was no existing study that applied deep learning for automatic analysis on heartbeat before the 2016 PhysioNet Computing in Cardiology Challenge. Now there are few attempts using deep learning models for classifying normal and abnormal heart sounds, which we describe next.

A deep learning based approach was used in \cite{rubin2017recognizing} for automatic recognition of abnormal heartbeat using a deep Convolutional Neural Network (CNN).
The authors computed a two-dimensional heat map from one-dimensional time series of PCG signal with the overlapping segment length of $T=3$ seconds and used for training and validation of the model. They achieved the highest specificity score $0.9521$ as compared to all entries made in PhysioNet Computing in Cardiology challenge but their sensitivity and accuracy scores were low: 0.7278 and 0.8399, respectively.  

A fully connected neural network (NN) consisting of 15 hidden layers was used in \cite{nassralla2017classification} for the classification of PCG signals. The authors achieved the recognition rate of $80\%$ with the specificity of 82 and poor sensitivity of 63. Potes et al. \cite{potes2016ensemble} used an ensemble of AdaBoost and the CNN classifiers to classify normal/abnormal heartbeats. This ensemble approach achieved the highest score in the among others and achieved \textit{rank one} in the competition with the specificity, sensitivity, and overall score of 0.7781, 0.9424, and 0.8602, respectively. Other approaches in this challenge were based on the classical machine learning based classifiers. Interestingly, none of the above studies have attempted RNN, which is, however, the most powerful model for time series data. 


RNN architectures such as Long Short-term Memory (LSTM) and Gated Recurrent Units (GRU) have not been used for PCG analysis but they have achieved state-of-the-art performances in various other applications with sequential data including speech recognition \cite{sutskever2014sequence,graves2013speech}, machine translation \cite{cho2014learning,bahdanau2014neural}, and emotion detection \cite{wollmer2008abandoning}. 
Bidirectional Long Short-Term Memory (BLSTM) \cite{schuster1997bidirectional}, which processes the information in both directions with two different LSTM layers show even better performance than LSTM and other conventional RNNs in phoneme classification and recognition \cite{graves2005bidirectional}. Moreover, Deep BLSTMs with the objective function of Connectionist Temporal Classification (CTC) are very effective in optimizing word error rate in speech recognition system with no prior linguistic or lexicon information \cite{graves2014towards}. Bidirectional Gated Recurrent Units (BiGRUs) have also been used for audio processing. They achieved a significant improvement in the performance of sound events detection from the recording of real-life, by capturing their temporal boundaries \cite{lu2017bidirectional}.



\section{Proposed Approach and RNN Models}
\label{sec:approach}
\begin{figure}[!ht]
\centering
\centerline{\includegraphics[width=.5\textwidth]{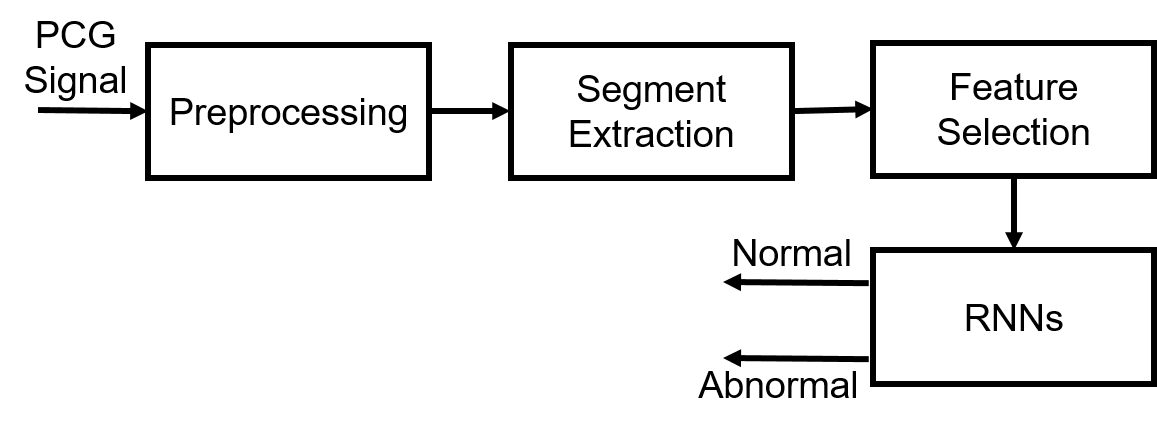}}
\caption{Block diagram of proposed approach}
\label{bl}
\end{figure}

Our proposed approach for heart sound classification using RNNs is depicted in Figure \ref{bl}. The heart sound is first prepossessed for first  and  second  heart  sounds (S1 and S2, respectively) detection and segmented into smaller chunks. Features extracted from these segments are given to the RNNs for classification, which classifies these into normal or abnormal. 

\subsection{Recurrent Neural Networks (RNNs)}

\begin{figure*}[!ht]%
\centering
\begin{subfigure}{0.31\linewidth}
\includegraphics[width=\linewidth]{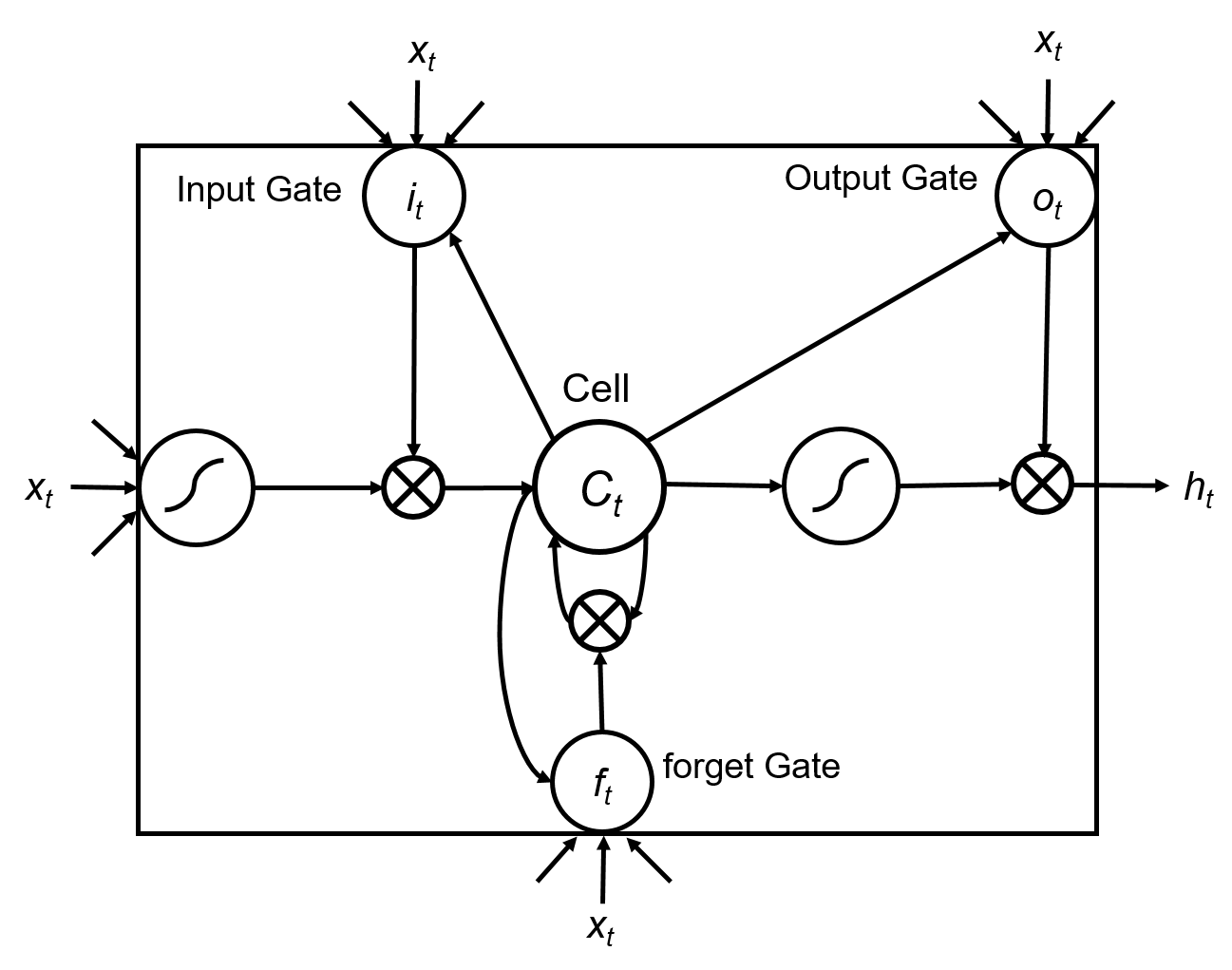}%
\captionsetup{justification=centering}
\caption{}%
\label{LSTM}%
\end{subfigure}
\begin{subfigure}{0.29\linewidth}
\includegraphics[width=\linewidth]{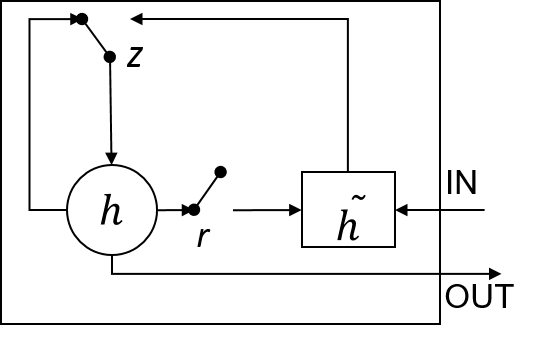}%
\captionsetup{justification=centering}
\caption{} %
\label{GRU}%
\end{subfigure}%
\begin{subfigure}{0.38\linewidth}
\includegraphics[width=\linewidth]{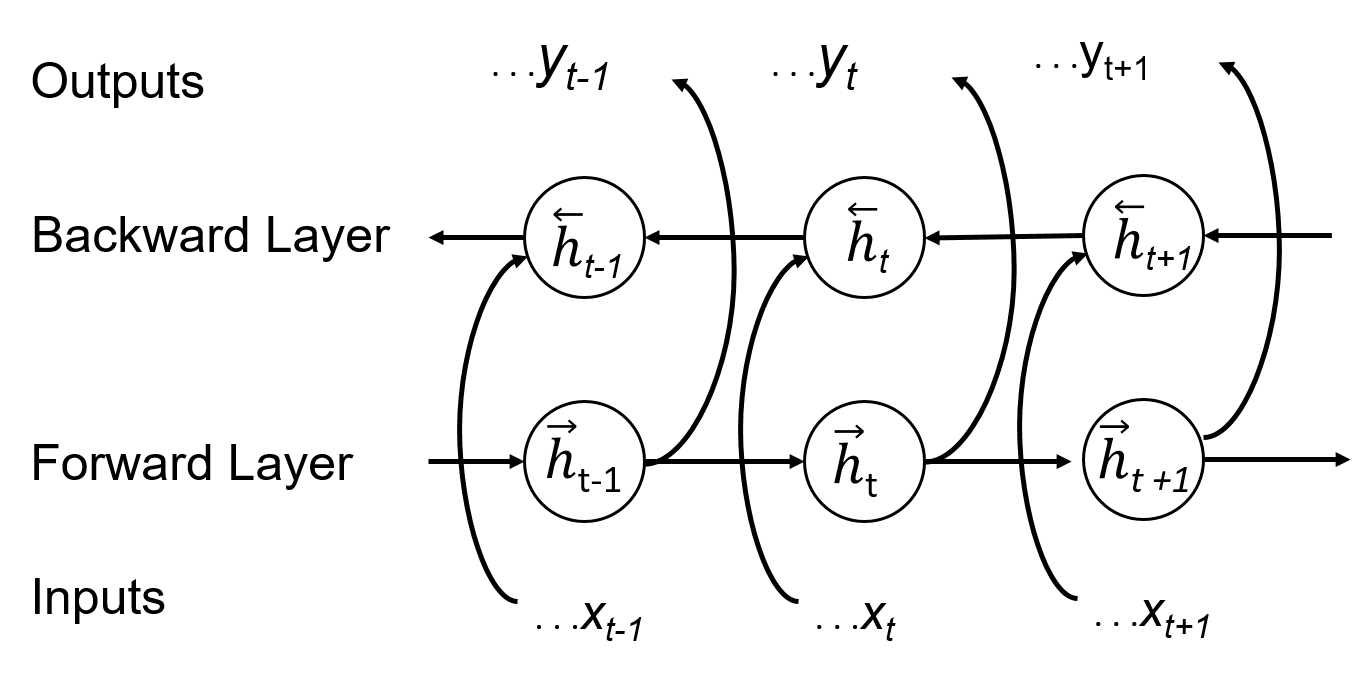}%
\captionsetup{justification=centering}
\caption{}%
\label{Bls}%
\end{subfigure}
\caption{Graphical representation of (\ref{LSTM}) LSTM memory cell; (\ref{GRU}) Gated Recurrent Unit (GRU); (\ref{Bls}) bidirectional LSTM.} 
\label{Graphic}
\end{figure*}

Recurrent neural networks (RNNs) are specialized to process sequences, unlike the CNNs, which are specialized for grid-like structures, such as, images. It takes an input sequence 
$x ({t})=(x_{1},.....,x_{T})$ and at the current time step $t$ calculates the hidden state or memory of the network $h_t$ using previous hidden state $h_{t-1}$ and the input $x_{t}$. The outputs from RNN are projected to the number of classes. A softmax function is used to project the output vector into the probability vector having values in [0,1]. When RNNs are used for abnormal heartbeat detection, its end layer is projected to the number of classes (normal and abnormal). Given a series of heart signal $x (t)=(x_{1},.....,x_{T})$, an RNN classifier learn to predict hypotheses $\hat{y}$ of true label $y$. 
The standard equations for RNN is given below:   
\begin{equation}
h_{t}= \textit{H}(W_{xh}x_{t}+W_{hh}h_{t-1}+b_{h})
\end{equation}
\begin{equation}
y_{t}= (W_{xh}x_{t}+b_{y})
\end{equation}
where $W$ terms are the weight matrices (i.e., $W_{xh}$ is weight matrix of an input-hidden layer), $b$ is the bias vector and $\textit{H}$ denotes the hidden layer function.


\subsubsection{Long Short-Term Memory (LSTM) Units}

The LSTM \cite{hochreiter1997long} network is a special type of RNN that 
consists of a recurrent memory block to store representations  
for extended time intervals. A memory block consists of three gates: input, output and forget gate. These multiplicative gates learn to control the constant error flow within each memory cell. The memory cell decides what to store, and when to enable reads, writes and erasures of information. Graphical representation of LSTM memory cell is shown in Figure \ref{LSTM}.

When features from a sequence of heart signal are given to the network, each LSTM unit holds a memory $c_{t}$ at a specific time $t$. The activation function is given by:
\begin{equation}
h_{t}=o_{t}\tanh(c_{t})
\end{equation}
The output gate $o_{t}$ modulates the memory content and calculated by:
\begin{equation}
o_{t}=\sigma\big(W_{xo}x_{t}+W_{ho}h_{t-1}+W_{co}c_{t}+b_{o}\big)
\end{equation}
The forget gate $f_{t}$  control the memory in the network and update it by forgetting the existing memory  $c_{t}$ with the incoming information.
\begin{equation}
f_{t}=\sigma\big(W_{xf}x_{t}+W_{hf}h_{t-1}+W_{cf}c_{t}+b_{f}\big)
\end{equation}
The extent of incoming information is controlled by the input gate $i_{t}$
\begin{equation}
i_{t}=\sigma\big(W_{if}x_{t}+W_{hi}h_{t-1}+W_{ci}c_{t}+b_{i}\big)
\end{equation}
The existing memory in the network is finally updated by the following equation under the control of these three gates.
\begin{equation}
c_{t}=f_{t}c_{t-1}+i_{t}\tanh\big(W_{xc}x_{t}+W_{hc}h_{t-1}+b_{c}\big)
\end{equation}

\subsubsection{Gated Recurrent Units (GRUs)}
The Gated Recurrent Unit (GRU) \cite{cho2014learning} is a slightly simplified version of LSTM that combines the input and forget gates into a single gate known as update gate. GRU architecture has an additional reset gate as compared to LSTM (see Figure \ref{GRU}). 
GRUs control the information flow from the previous activation while computing candidate activation but unlike LSTM do not control the amount of incoming memory using the forget gate.  








\subsection{Bidirectional Recurrent Neural Networks }
One shortcoming of standard RNNs is that they can only use previous information for making decisions. Bidirectional RNNs  \cite{schuster1997bidirectional} using LSTM units or GRUs can process the information both in the forward and backward direction which enable them to exploit future context. 
This is achieved by computing the hidden sequence both in forward $\overrightarrow{h}$ and backward direction $\overleftarrow{h}$,  and updating the output layer using backward layer from time step $t=T$ to $1$ and forward layer from $t=1$ to $T$ (see Figure \ref{Bls}). 





Similarly, a bidirectional GRU (BiGRU) exploits the full use of contextual information and produces two sequences $[h_{1}^f,h_{2}^f,.....,h_{T}^f]$ and $[h_{1}^b,h_{2}^b,.....,h_{T}^b]$ by processing the information both in forward and backward directions, respectively.   These two sequences are concatenated at the output by the following equation:

\begin{equation}
   \overleftrightarrow{GRU}(X)=|h_{1}^{f}||h_{T}^{b},...,h_{T}^{f}||h_{1}^{b}|.
\end{equation}
Here, the $\overleftrightarrow{GRU}(X)$ term represents the full output of BiGRU produced by concatenating each state in forward direction $h_{i}^{f}$ and backward direction $h_{(T-i+1)}^{b}$ at step $i$ given the input $X$.
In this paper, we use both RNNs and bidirectional RNNs and compare their performances on abnormal heartbeat detection. 

\section{Experimental Procedure}

The performance of different RNN models is evaluated on the publicly available datasets. The details of the datasets and experimental procedure are presented in this section. 

\subsection{Database Description}

To evaluate the proposed methodology, the largest dataset provided at the ``Physionet Challenge 2016'' \cite{liu2016open} has been used.
The Physionet dataset consists of six databases (A through F) containing a total of 3240 raw heart sound recordings. These recordings were independently collected by different research teams using heterogeneous sensing equipment from different countries both in clinical
and nonclinical (i.e., home visits) settings. The dataset contains both clean and noisy heart sound recordings. The recordings were collected both from healthy subjects and patients with a variety of heart conditions, especially coronary artery disease and heart valve disease. The subjects were from different age groups including children, adults and elderly. The length of heart sound recordings varied from 5 seconds to just over 120 seconds. 
For our experiments, we use all six databases containing normal and abnormal heart sound recordings. 



\subsection{Preprocessing of Heart Sound}


The heart sound recorded by electronic stethoscope often has background noise. The preprocessing of heart sound is an essential and crucial step for automatic analysis of heartbeat recordings. It reveals the inherent physiological structure of the heart signal by detecting the abnormalities in the meaningful regions of PCG signal and allows for the automatic recognition of pathological events. The detection of the exact locations of the first  and second heart sounds (i.e., S1 and S2) within PCG is known as the segmentation process. The main goal of this process is to ensure that incoming heartbeats are properly aligned before 
their classification as it significantly improves the recognition scores \cite{deperlioglu2018classification}. 

In this paper, we used state-of-the-art method Logistic Regression-Hidden Semi-Markov Models (HSMM) for identification of heart states proposed by Springer et al. \cite{springer2016logistic}. This method uses LR-derived emission or observation probability estimates and provides significantly improved results as compared to the previous approaches based on the Gaussian or Gamma distributions \cite{schmidt2010segmentation,hughes2006probabilistic}. 

The working of Logistic Regression-HSMM is similar to SVM based emission probabilities \cite{marzbanrad2014automated} and it allows for greater discrimination between different states. Logistic regression is a binary classifier that maps the feature space or predictor variables to the binary response variables by using a logistic function. The logistic function $\sigma(a)$ is defined as: 
\begin{equation}
    \sigma(a)=\frac{1}{1+\exp{(-a)}}
\end{equation}
The probability of a state or class given the input observations $O_{t}$ can be defined using the logistic function:
\begin{equation}
    P\big[q_{t}=\xi|O_{t}\big]=\sigma(w^{'}O_{t}).
\end{equation}
The term $w$ represents the weights of the model that are applied to each observation or input features. The model is trained iteratively and re-weighted least squares on the training data. For one-vs-all logistic regression, the probability of each observation given the state $b_{j}(O_{t}|\xi_{j})$ is found by using Bayes'rule:
\begin{equation}
    b_{j}(O_{t})=P\big[O_{t}|q_{t}=\xi\big]=\frac{P\big[q_{t}=\xi|O_{t}\big] \times P\big(O_{t}\big)}{P\big(\xi_{j}\big)}.
\end{equation}
The $P(O_{t})$ is calculated from a multivariate normal distribution of the entire training data and $P(\xi_{j})$ is the initial state probability distribution. 

The Logistic Regression-HSMM algorithm use the combination of four type of features: Homomorphic envelope, Hilbert envelope, Wavelet envelope and Power spectral density envelope. The details of these features can be seen in \cite{springer2016logistic}. The overall PCG recordings are given to the model for accurate detection of most probable states (i.e., S1 and S2). 

\begin{figure}[!ht]
\centering
\centerline{\includegraphics[width=0.48\textwidth]{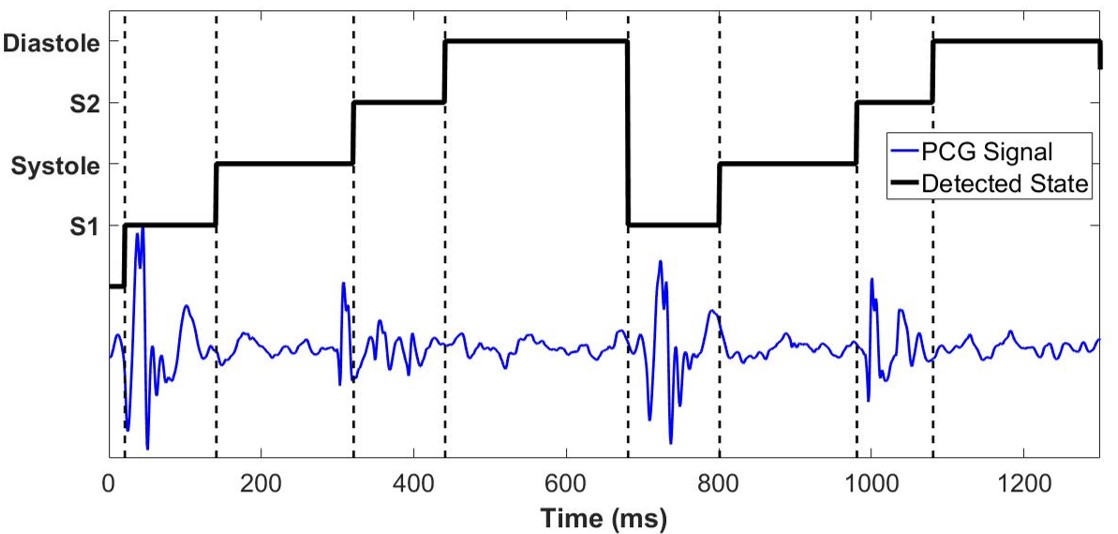}}
\caption{Four states (S1, S2, systole, and diastole) of the heart cycle using Logistic Regression-HSMM}
\label{detected}
\end{figure}

Figure \ref{detected} shows the detected four states (i.e., S1,  S2,  systole, and asystole) of two heart cycles. Note that, generally, it is called as S1 and S2 detection, although it detects all the four states: S1, S2, systole, and asystole. The blue line is for the heart signal and the black line shows the detected states using Logistic Regression-HSMM algorithm.

\begin{figure*}[!t]%
\centering
\begin{subfigure}{0.5\linewidth}
\includegraphics[width=\linewidth]{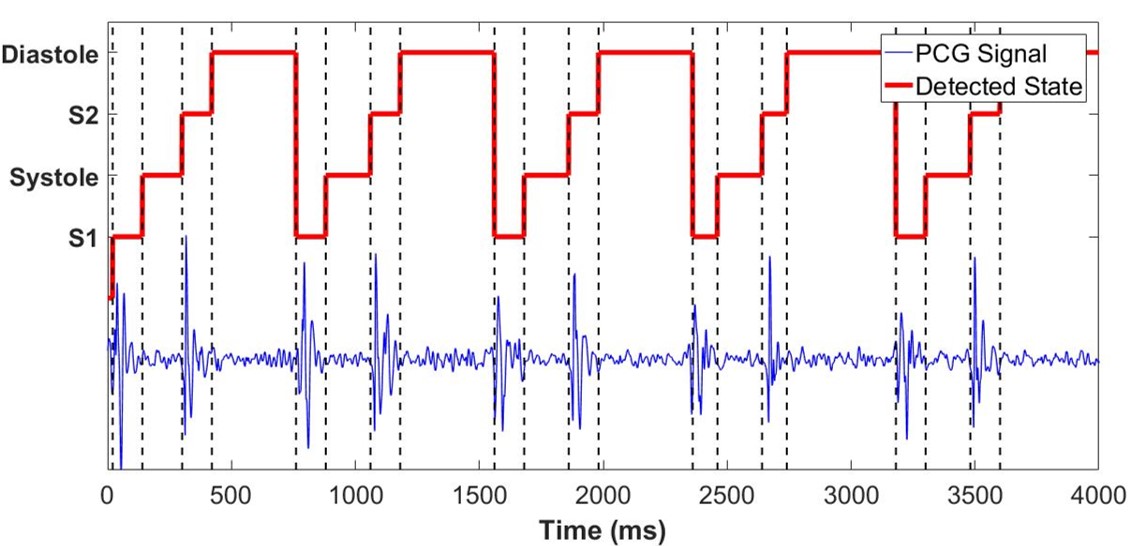}%
\captionsetup{justification=centering}
\caption{}%
\label{normal}%
\end{subfigure}
\begin{subfigure}{0.5\linewidth}
\includegraphics[width=\linewidth]{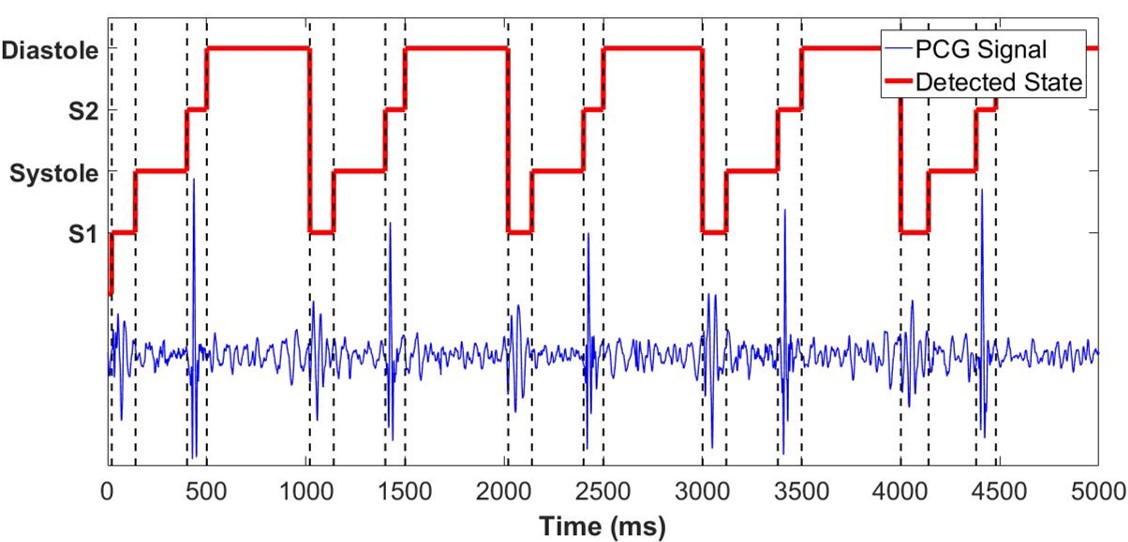}%
\captionsetup{justification=centering}
\caption{} %
\label{abnormal}%
\end{subfigure}%
\caption{Extracted segments of five heart cycles using (\ref{normal}) Normal and (\ref{abnormal}) Abnormal heartbeats.  {\bf Abnormal heart cycles have longer duration.}}
\label{segmentsh}
\end{figure*}

\subsection{Segment Extraction}
After detecting the position of heart states, we segmented the overall PCG waveform into shorter instances by locating the beginning of each heartbeat. This is because the number of audio recordings (i.e., 3240) of heart signal is not adequate to evaluate RNNs for designing a robust system. Segment extraction was also used in previous studies to divide the overall heart sound in smaller chunks. For instance, Rubin et al. \cite{rubin2017recognizing} used a segment of $T=3$ seconds for training and validation of CNN. In this paper, we segmented the heart signal into three sequences of heart cycles: 2, 5, and 8. These segments contain enough information and at the same time small enough to generate many samples for training the model. Figure \ref{normal} \& \ref{abnormal} shows the five cycles of normal and abnormal heart sound, respectively. The abnormal heart sound is different from the normal one in temporal context. It has heart cycle states of longer duration in the segment. 

\subsection{Feature Selection}
The extraction and selection of more relevant parameters from PCG signals are crucial tasks. They significantly affect the recognition
performance of the model.  In this paper we used Mel-frequency cepstral coefficients (MFCCs) \cite{davis1980comparison} to represent PCG signal in compact representation. MFCCs are used almost in every study on automatic heart sound classification (for example,  \cite{ortiz2016heart,potes2016ensemble,zabihi2016heart,rubin2016classifying}) due to their effectiveness in speech analysis. 
We compute MFCCs from 25ms of the window with a step size of 10ms. We select the first 13 MFCCs for compact representation of PCG signal as a large feature space does not always improve the recognition rate of the model \cite{rana2016gated}.

\subsection{Model Parameters}


We built our models using Keras \cite{chollet2015keras} with a TensorFlow backend \cite{abadi2016tensorflow}. In order to find the best models' structure, we evaluated a different number of gated layers from one to four. A smaller model with only one LSTM or GRU layer was not performing well on this task. Experiments with a larger number of gated layers and dense layers also failed to give improvements in performance. This might be due to overfitting problem. We got the best classification results using 2 gated layers for both LSTM and GRU models. Therefore, our LSTM and BLSTM models consist of two LSTM layer with $\tanh$ function \cite{glorot2010understanding} as activation.  For each heartbeat, the outputs of LSTM or BLSTM layers were given to the dense layer and the outputs of this layer were given to the softmax layer for classification (refer to Section \ref{sec:approach}).  


We trained all these models using the training portion of the dataset and development portion of the data was used for hyper-parameter selection. We started training the network with a learning rate of 0.002 and learning rate was halved after every 5 epochs if the classification rate of the model did not improve on the validation set. This process continued until the learning rate reached below 0.00001 or until the maximum epochs i.e., 100 were reached. We used batch normalization after the dense layer for normalization of learned distribution to improve the training efficiency \cite{ioffe2015batch}. In order to accommodate the effect of initialization,  we repeated each hyperparameter combination for three times and used the averaged prediction for validation and testing.

\section{Results and Discussion}
\begin{figure}[!ht]
\centering
\centerline{\includegraphics[width=.45\textwidth]{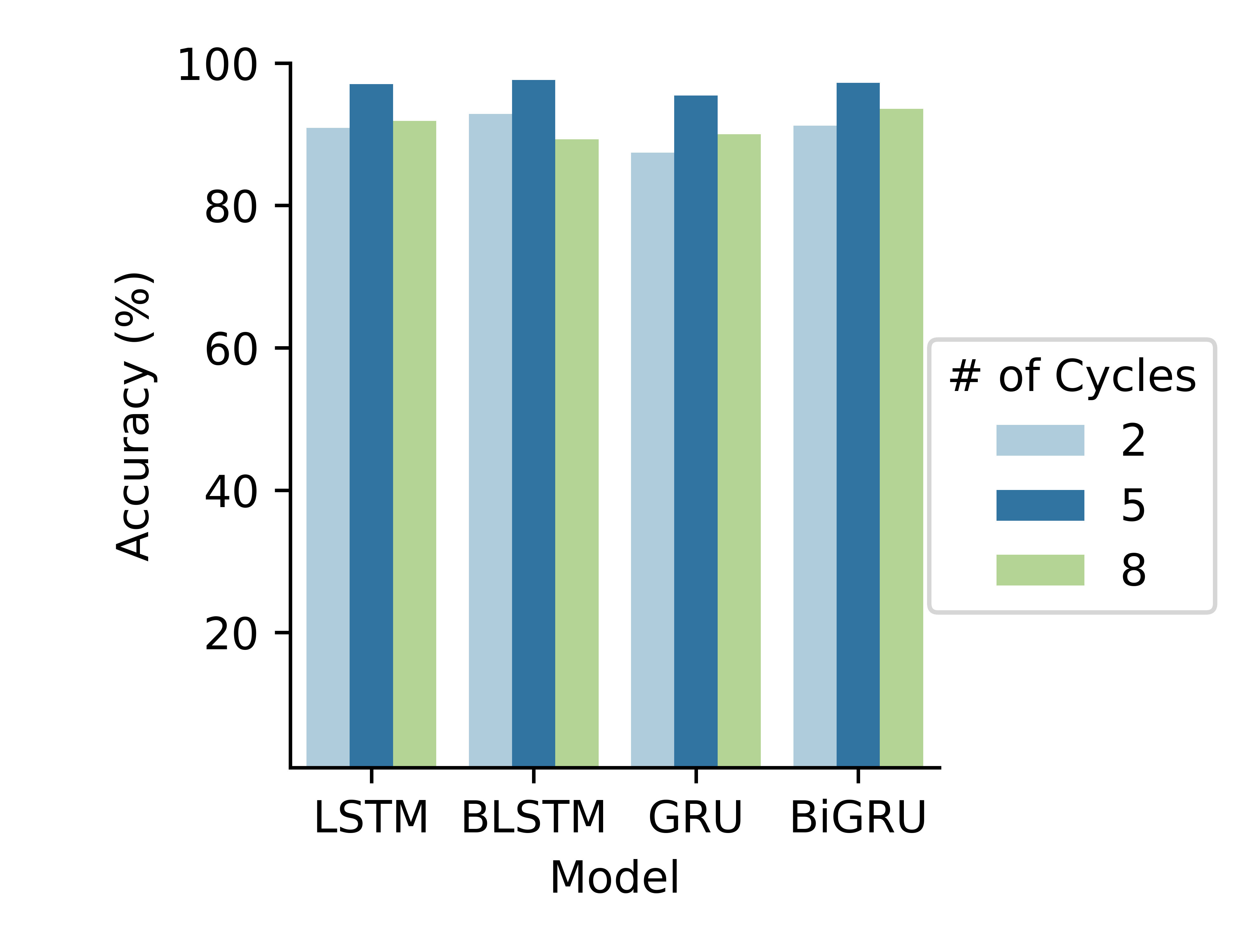}}
\caption{Accuracy using different number of heart cycles, where RNNs give best results using 5 cycles.}
\label{Cycles}
\end{figure}

The overall dataset of PhysioNet Computing in Cardiology Challenge consisted of eight heart sound databases gathered from seven different countries. There is a total of 3240 publicly available heartbeat recordings. We detected heart states in each PCG waveform and segmented these signals into smaller chunks containing exact five heart cycles. MFCCs were computed from these chunks and both models were trained on 75\% of data, 15\% data was used for validation and remaining 10\% of data was used for testing. As each model was trained on MFCC computed from the smaller segment, to predict the score for full instance, an averaging was performed on posterior probabilities of the respective chunks. In this work, the accuracy of RNNs has been tested on three different cycles: 2, 5, and 8 in a heart signal (see Figure \ref{Cycles}). All RNN models consistently performed well on 5 cycles. Therefore, onward we mention results using 5 cycles.

\begin{figure*}[!ht]%
\centering
\begin{subfigure}{0.5\linewidth}
\includegraphics[width=\linewidth]{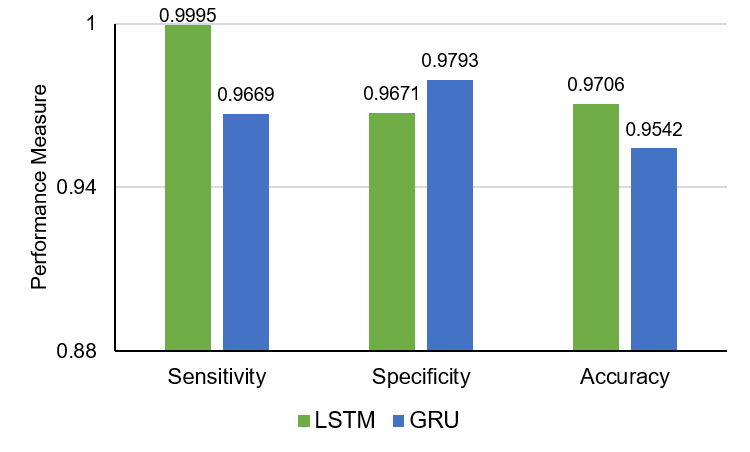}%
\captionsetup{justification=centering}
\caption{}%
\label{fig:a}%
\end{subfigure}
\begin{subfigure}{0.5\linewidth}
\includegraphics[width=\linewidth]{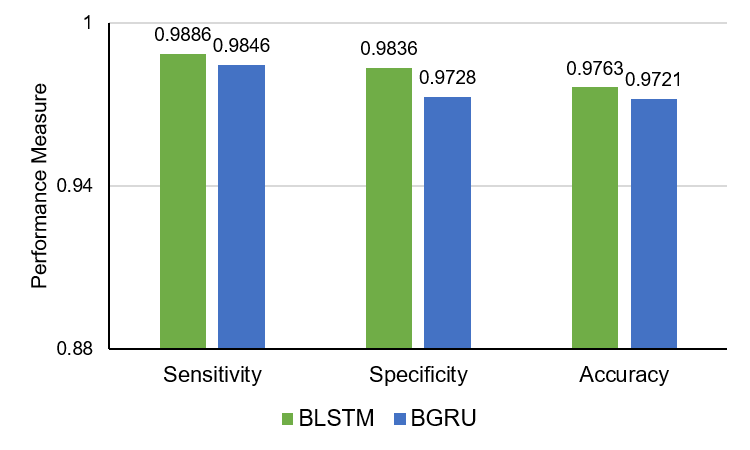}%
\captionsetup{justification=centering}
\caption{} %
\label{fig:b}%
\end{subfigure}%
 \caption{Performance comparison of RNNs on heartbeat classification: (\ref{fig:a}) LSTM vs GRU and (\ref{fig:b}) BLSTM vs BiGRU.}
\label{segments}
\end{figure*}

\subsection{Baseline models}
In order to compare the performance of RNNs, we selected three classic powerful classifiers: Logistic Regression (LR), Support Vector Machines (SVM), and Random Forest (RF) as baseline models. We trained these models on MFCCs computed from 5 heart cycles. In SVM, we tested three kernels: linear, polynomial, and radial basis function (RBF) to obtain the best classification results.  Hyperparameters of these models were selected using validation data. Table \ref{base} shows the best results obtained by these models and presents a comparison with RNNs. We observe RNNs significantly outperform all three models in every performance measures. 

\begin{table}[!ht]
\centering
\caption{Comparison of SVM and LR with RNN}
\label{base}
\begin{tabular}{|c|c|c|c|}
\hline
\textbf{Model}       & \textbf{Sensitivity} & \textbf{Specificity} & \textbf{Accuracy} \\ \hline
SVM (best)  & 0.8259      & 0.8324      & 0.8291   \\ \hline
LR (best)   & 0.7121      & 0.6879      & 0.6991   \\ \hline
RF (best)   & 0.6901      & 0.6850      & 0.6861   \\ \hline
RNNs (best) & {\bf 0.9886}      & {\bf 0.9836}      & {\bf 0.9763}   \\ \hline
\end{tabular}
\end{table}


\subsection{Comparison with non-recurrent deep models} 

We compare the performance of RNNs with most recent studies on PCG classification using DNNs in Table \ref{table: Com}. 
\begin{table}[!h]
\centering
\caption{Comparison of results with previous attempts.}
\begin{tabular}{|m{2cm}|m{1.3cm}|m{1.08cm}|m{1.08cm}|m{1cm}|}
\hline
\textbf{Author (Year)}
&\textbf{Approach}
&\textbf{\scriptsize{Sensitivity}}
&\textbf{\scriptsize{Specificity}}
&\textbf{\scriptsize{Accuracy}}
\\ \hline
\begin{tabular}[c]{@{}l@{}}Potes et al. \cite{potes2016ensemble}\\ (2016)\end{tabular}
&\begin{tabular}[c]{@{}l@{}}AdaBoost \\and CNN \end{tabular}
&\begin{tabular}[c]{@{}l@{}}0.9424\end{tabular}
&\begin{tabular}[c]{@{}l@{}}0.7781\end{tabular}
&\begin{tabular}[c]{@{}l@{}}0.8602\end{tabular}
\\\hline
\begin{tabular}[c]{@{}l@{}}Tschannen et al. \\\cite{tschannen2016heart} (2016)\end{tabular}
&\begin{tabular}[c]{@{}l@{}}Wavelet-\\based CNN \end{tabular}
&\begin{tabular}[c]{@{}l@{}}0.855\end{tabular}
&\begin{tabular}[c]{@{}l@{}}0.859\end{tabular}
&\begin{tabular}[c]{@{}l@{}}0.828\end{tabular}
\\\hline
\begin{tabular}[c]{@{}l@{}}Rubin et al. \cite{rubin2017recognizing}\\(2017)\end{tabular}
&\begin{tabular}[c]{@{}l@{}}CNN \end{tabular}
&\begin{tabular}[c]{@{}l@{}}0.7278\end{tabular}
&\begin{tabular}[c]{@{}l@{}}0.9521\end{tabular}
&\begin{tabular}[c]{@{}l@{}}0.8399\end{tabular}
\\\hline
\begin{tabular}[c]{@{}l@{}}Nassralla at al. \\\cite{nassralla2017classification} (2017)\end{tabular}
 &\begin{tabular}[c]{@{}l@{}}DNNs \end{tabular}
&\begin{tabular}[c]{@{}l@{}}0.63\end{tabular}
&\begin{tabular}[c]{@{}l@{}}0.82\end{tabular}
&\begin{tabular}[c]{@{}l@{}}0.80\end{tabular}\\ \hline
\begin{tabular}[c]{@{}l@{}}Dominguez at al. \\\cite{dominguez2018deep} (2018)\end{tabular}
 &\begin{tabular}[c]{@{}l@{}}Modified\\ AlexNe \end{tabular}
&\begin{tabular}[c]{@{}l@{}}0.9512\end{tabular}
&\begin{tabular}[c]{@{}l@{}}0.9320\end{tabular}
&\begin{tabular}[c]{@{}l@{}}0.9416\end{tabular}\\ \hline
&\begin{tabular}[c]{@{}l@{}}LSTM\end{tabular}
&\begin{tabular}[c]{@{}l@{}}\textbf{0.9995}\end{tabular}
&\begin{tabular}[c]{@{}l@{}}0.9671\end{tabular}
&\begin{tabular}[c]{@{}l@{}}0.9706\end{tabular}\\\cline{2-5}
\begin{tabular}[c]{@{}l@{}}Our Study (2018)\end{tabular}
&\begin{tabular}[c]{@{}l@{}}BLSTM\end{tabular}
&\begin{tabular}[c]{@{}l@{}}0.9886\end{tabular}
&\begin{tabular}[c]{@{}l@{}}\textbf{0.9836}\end{tabular}
&\begin{tabular}[c]{@{}l@{}}\textbf{0.9763}\end{tabular}\\\cline{2-5}
 &\begin{tabular}[c]{@{}l@{}}GRU\end{tabular}
&\begin{tabular}[c]{@{}l@{}}0.9669\end{tabular}
&\begin{tabular}[c]{@{}l@{}}0.9793\end{tabular}
&\begin{tabular}[c]{@{}l@{}}0.9542\end{tabular}\\\cline{2-5}
 &\begin{tabular}[c]{@{}l@{}}BiGRU\end{tabular}
&\begin{tabular}[c]{@{}l@{}}0.9846\end{tabular}
&\begin{tabular}[c]{@{}l@{}}0.9728\end{tabular}
&\begin{tabular}[c]{@{}l@{}}0.9721\end{tabular}\\ \hline
\end{tabular}
\label{table: Com}
\end{table}
It can be clearly noticed that RNNs outperform all deep learning models used on heart sound classification with a significant improvement. Worth mentioning, heart sound classification using the ensemble of AdaBoost and CNN achieved ``Rank 1'' in  PhysioNet Computing in Cardiology Challenge 2016 with the best recognition rate. In our approach, RNNs using LSTM and even with GRUs have achieved better recognition rate than that in every performance measures.

\subsection{Performance Comparison of RNNs}

In this study, our evaluation focused on the sequence modeling of heart sound using RNNs. We explored the performance of different state-of-the-art RNNs for this task. Based on the experimental results, it can be highlighted that the different RNNs have achieved comparable performances. 
{However, out of all models, BLSTMs perform consistently  well and is preferred for heartbeat classification using PCG.} Another important finding, despite having a simpler architecture compared to LSTM, the performance of GRU is also promising on PCG data.




\section{Conclusion}
In this paper, we conduct an empirical study on abnormal heartbeat detection using PCG signal and demonstrate that Recurrent neural networks (RNNs) produce promising results. In particular, the results are significantly better than the conventional deep learning models. RNNs have the most important architectures to capture the temporal statistics and dynamics in the sequence of heartbeats more efficiently as compared to the other popular DNNs like CNN. 
Noteworthy, the RNN models significantly outperform the AdaBoost-CNN model which was placed Rank 1 in  the PhysioNet  Computing  in  Cardiology  Challenge  2016. In our future studies, we aim to produce further empirical evidences of why RNNs are promising for heartbeat classification. 


\bibliographystyle{IEEEtran}
\bibliography{Ref}

\end{document}